# Phoneme Recognition Using Acoustic Events[*]

*Kai Hübener*

University of Hamburg, Germany
Computer Science Dept.

*Julie Carson–Berndsen*

University of Bielefeld, Germany
Faculty for Linguistics and Literary Studies

## Abstract

This paper presents a new approach to phoneme recognition using nonsequential sub–phoneme units. These units are called acoustic events and are phonologically meaningful as well as recognizable from speech signals.

Acoustic events form a phonologically incomplete representation as compared to distinctive features. This problem may partly be overcome by incorporating phonological constraints. Currently, 24 binary events describing manner and place of articulation, vowel quality and voicing are used to recognize all German phonemes.

Phoneme recognition in this paradigm consists of two steps: After the acoustic events have been determined from the speech signal, a phonological parser is used to generate syllable and phoneme hypotheses from the event lattice. Results obtained on a speaker–dependent corpus are presented.

## 1 Introduction

It is well known that not phonemes but distinctive features constitute the smallest phonologically meaningful units in spoken language (?). In turn each phoneme can be uniquely identified by a subset of these features, e. g. the phoneme /d/ is uniquely specified by the features [+cons, -voc, +voiced, +anterior, +coronal]. About 20 distinctive features are needed to specify all phonemes of any given language. Using distinctive features for phoneme recognition has several advantages: They allow a better modelling of phonological phenoma such as assimilation and coarticulation, and they are less abstract than allophonic units or phonemes and thus acoustic correlates should be found more easily as compared to allophones.

Although distinctive features are traditionally based on minimal pairs of phonemes, they are closely related to articulatory movements or gestures. However, this makes their acoustic correlates highly context–dependent and thus hard or even impossible to detect in speech signals. Even worse, directly observable correlates may not exist at all for some features. Nevertheless, some results for direct recognition of some distinctive features have been reported recently (?; ?).

However, although phonologically meaningful, distinctive features do not have any explicit notion of temporality which is clearly required for recognition purposes. For this reason a new approach is taken here whereby top-down constraints on temporal relations between events are defined at the phonological level allowing different synchronisation functions for each event type. This caters explicitly for the notion of overlap and precedence and avoids a segmentation of the speech signal using a single synchronisation function as in direct phoneme recognition.

The rest of this paper is organized as follows: In Section ?? the concept of acoustic events is introduced. The recognition of these events from speech signals is described in Section ??. Section ?? introduces the phonological parser used to generate syllable and phoneme hypotheses from acoustic event lattices. In Section ?? the motivation for the use of events in phonological parsing is presented and the declarative phonological knowledge base is described in Section ??. The recognition of phonemes and syllables is presented in Section ?? together with some initial results.

## 2 Acoustic Events

Since acoustic counterparts for important distinctive features e. g. [consonant] are not easily found, we define the concept of acoustic events which are phonologically meaningful as well as recognizable from speech signals (?). Acoustic events are nonsequential units, i. e. they may overlap in time. The phoneme /d/ for instance is characterized by the acoustic events [sh, ap, op]. However, due to the latter criterion, acoustic events form a phonologically underspecified representation, i. e. some phonemes may not be uniquely identifiable given the acoustic events. Currently, a set of 24 different acoustic events shown in Table ?? is used to describe manner and place of articulation, vowel quality and voicing.

| fr | fricative | vg | rounded vowel |
| gh | noisy | vm | mid vowel |
| na | nasal | vt | tense vowel |
| op | occlusion | vr | rounded vowel |
| pa | pause | vu | unrounded vowel |
| sh | voiced | lb | labial |
| tv | transient | ap | apical |
| va | a–like vowel | po | palato |
| vd | dark vowel | pl | palatal |
| vh | light vowel | ve | velar |
| vo | vowel | gl | glottal |
| vz | central vowel | la | lateral |

Table 1: Acoustic Events

These events were found by repeatedly optimizing the mapping from phonemes to sets of acoustic events. This includes the generation of appropriately labelled training material, training detectors for the new events, running the recognizer and analyzing the recognition results and errors.

## 3 Event Recognition

Acoustic events are binary–valued: They are either present or absent. The presence of an event may require the absence of others, e. g. [pa] and [sh] are mutually exclusive. These relationships are mostly articulatory and phonological constraints. Consequently, they should modelled at the phonological level and not at the acoustic level in order to keep phonological knowledge apart from acoustic knowledge. Nevertheless, allowing feedback from the phonological level to the acoustic event recognizer is possible at this point. Currently such phonological constraints at the acoustic level are ignored allowing greater flexibility when experimenting with different event sets and easy expansion of the acoustic event set.

---
[*]This research was supported by the German Minister of Research and Technology (BMFT) under grants 01IV101A/0 and 01IV10B2. Views and conclusions contained in this paper are those of the authors and should not be interpreted as representing the official opinion or policy of BMFT or the German Government.

To recognize an event the speech signal is sampled at 16 kHz and blocked into 30 ms frames which overlap by 20 ms. Each frame is parametrized using five cepstrally smoothed PLP coefficients (?). Additionally, log energy and regression coefficients are appended resulting in 13-element feature vectors. Alternatively, traditional mel-cepstral coefficients as well as RASTA-PLP coefficients were investigated but these led to inferior results. The feature vectors are classified using two quadratic Bayesian classifiers trained on presence and absence of an event. Adjacent frames are assumed to be statistically independent since including features from adjacent frames led to inferior performance as well as the inclusion of segmental boundary information (?).

To evaluate the performance of the acoustic event classifiers the recognizer was trained on 180 read utterances from a single speaker. Training labels were created automatically by translating aligned phonemic transcriptions obtained from a SCHMM recognizer into event labels. Frame-based event recognition rates between 77% and 98% were achieved on an independent test set of 20 utterances (Table ??). This clearly demonstrates the recognizability of acoustic events from speech signals. The event recognizer runs in real time on a Sparc10.

| Event | Occur | Corr. | False Alarms | | Misses | |
|---|---|---|---|---|---|---|
| | | | abs | rel | abs | rel |
| fr | 15.41 | 90.71 | 3.95 | 4.67 | 5.34 | 34.64 |
| gh | 11.58 | 94.25 | 3.14 | 3.55 | 2.62 | 22.59 |
| na | 10.57 | 92.02 | 1.39 | 1.56 | 6.59 | 62.37 |
| op | 9.12 | 90.77 | 0.15 | 0.16 | 9.08 | 99.57 |
| pa | 37.97 | 90.60 | 4.34 | 7.00 | 5.05 | 13.31 |
| sh | 44.18 | 89.59 | 4.20 | 7.53 | 6.21 | 14.05 |
| tv | 2.27 | 97.06 | 1.11 | 1.13 | 1.84 | 80.87 |
| va | 6.81 | 95.08 | 2.54 | 2.72 | 2.39 | 35.07 |
| vd | 13.18 | 93.58 | 3.48 | 4.01 | 2.94 | 22.31 |
| vh | 6.71 | 95.41 | 2.41 | 2.58 | 2.18 | 32.50 |
| vo | 25.22 | 89.16 | 7.18 | 9.61 | 3.65 | 14.48 |
| vz | 13.38 | 85.70 | 7.79 | 8.99 | 6.51 | 48.67 |
| vg | 7.22 | 92.42 | 3.56 | 3.84 | 4.02 | 55.60 |
| vm | 8.62 | 90.30 | 4.55 | 4.98 | 5.15 | 59.79 |
| vt | 9.39 | 93.45 | 3.42 | 3.78 | 3.13 | 33.33 |
| vr | 7.86 | 94.08 | 1.95 | 2.12 | 3.97 | 50.50 |
| vu | 17.37 | 88.17 | 7.33 | 8.87 | 4.50 | 25.91 |
| lb | 9.64 | 91.05 | 0.36 | 0.39 | 8.60 | 89.15 |
| ap | 17.05 | 85.13 | 5.58 | 6.72 | 9.30 | 54.51 |
| po | 0.72 | 99.47 | 0.27 | 0.27 | 0.27 | 36.99 |
| pl | 2.52 | 97.52 | 1.40 | 1.44 | 1.08 | 42.75 |
| ve | 5.31 | 95.10 | 0.77 | 0.81 | 4.14 | 77.88 |
| gl | 0.53 | 99.47 | 0.00 | 0.00 | 0.53 | 100.00 |
| la | 1.51 | 98.25 | 0.32 | 0.32 | 1.43 | 94.77 |
| uv | 0.84 | 99.12 | 0.04 | 0.04 | 0.84 | 100.00 |

Table 2: Event recognition rates

As can be seen from Table ?? acoustic events describing vowel quality and place of articulation are detected less reliably than other events. This may be explained by the absence of formant frequencies from the feature vector. The relatively poor performance of the silence detector is due to inconstent labeling of the test/training set since occlusions during plosives and pauses between words were not included in labelled material.

An attempt was made to smooth the detector outputs using a fixed-sized windows but this did not improve performance except in the case of the silence detector since detection of occlusions and inter-word silence was suppressed (?).

To demonstrate the phonological relevance of acoustic events two different event-based recognition schemes were tried, a stochastic approach and a knowledge-based approach.

For the stochastic approach, the frame-synchronous aposteriori probabilities for each event were stacked to form 24-element feature vectors which were used as observation vectors in a semicontinuous HMM recognizer. This led to a phoneme recognition rate of about 42% on the above test set which is comparable to (?). The recognition rate improved to 58% when phoneme bigrams were used.

For the knowledge-based approach, the acoustic event lattice from the event recognizer was analyzed using a phonological parser which is described in the following sections.

## 4 The Phonological Event Parser

The task of the phonological event parser is to construct syllable and phonological word hypotheses from the acoustic event lattice and, in doing so, to restrict the search space of other modules in the spoken language recognition system. For phonological parsing a flexible notion of compositionality is utilised in line with recent developments in multilinear (autosegmental) phonology (?; ?; ?). This approach is based on underspecified structures with 'autosegmental' tiers of parallel phonological events which avoids a rigid mapping from acoustic parameters to simple sequences of phoneme segments.

The phonological event parser imposes top-down contraints on the acoustic event lattice in the form of well-formedness constraints on permissible syllable structures, in this case for German. Assuming that at the acoustic event detector level all combinations of sounds are considered possible, phonotactic constraints are applied in order to reduce the search space to cover only those combinations which are permissible in the language. The phonotactic constraint system describes an autosegmental representation of phonological events and the temporal relations (overlap, precedence and inclusion) which exist between them. This is in line with results of work in articulatory phonology by Browman and Goldstein (?; ?) in connection with a parametric speech synthesis system where overlapping gestures from individual articulators all contribute to the realization of a particular utterance: differing degrees of overlap result in differing phonetic realizations. The notion of hidden or blended gestures in the case of fast speech where nothing is deleted or inserted allows a departure from rules describing phonological processes. In the analysis direction, independent acoustic events contribute different information which is relevant for the composition of phonological events.

## 5 The Projection Problem

The main motivation for the application of the event concept at the phonological level in spoken language recognition systems concerns the one of the major problems in the processing of speech, namely the projection problem at the phonetics/phonology interface (?) with respect to the compositionality and variability of speech. Sounds and words are realized with different degrees of coarticulation (overlap of properties) in different lexical, syntactic and phonostylistic contexts and thus a segmentation into phonemes alone is too rigid in order to capture all variants. Furthermore, the set of possible words in natural languages, analogous to the set of sentences, is infinite. In fact, even finite subsets of these sets e. g. the set of syllables, may be so large that a simple list is no longer tractable. This has so far proved to be an insuperable problem for the primarily concatenative word models of current speech recognition systems whether phoneme, demisyllable or word based. In linguistic terms, the projection problem refers to the predictive skill of the native speaker which allows the projection of a finite set of **actual** structures onto a (possibly infinite) set of **potential** structures. In order to go some way to solving the projection problem, a more flexible nonsegmental approach to spoken language recognition is chosen in the context of phonological parsing which incorporates the notion of compositionality by employing several sources of information simultaneously. Such an approach avoids a strict classification in terms of demi-syllables, phonemes, or even phones.

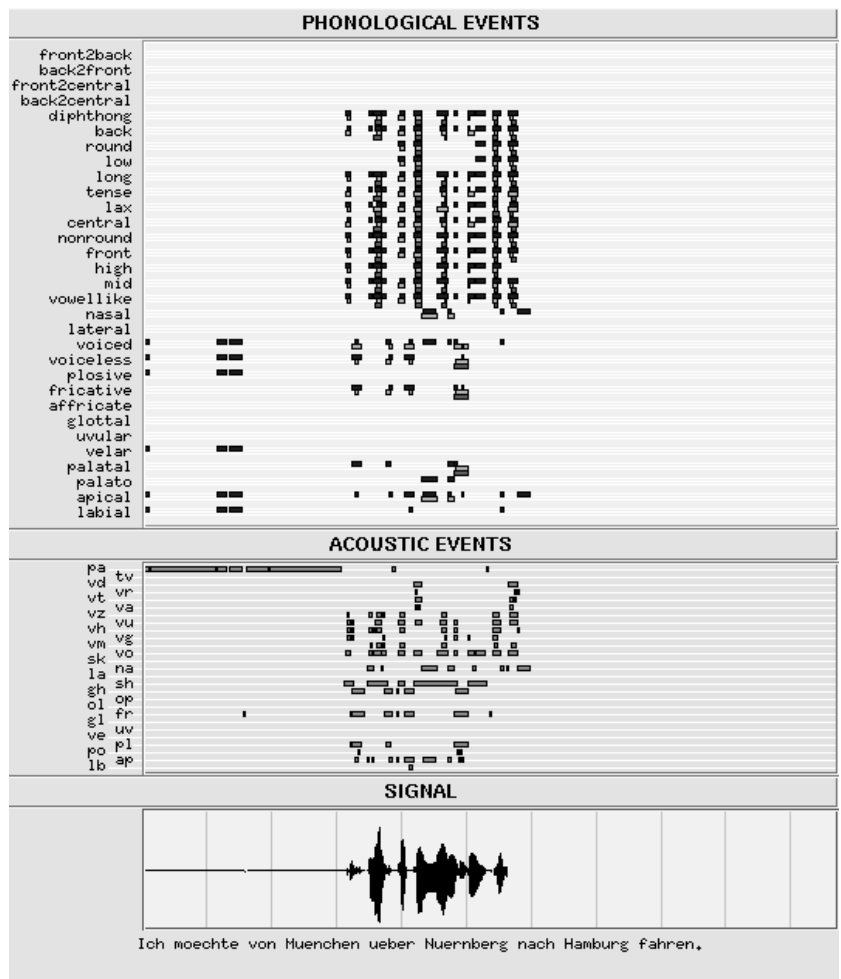

Figure 1: Incremental Mapping from Acoustic to Phonological Events

## 6  The Phonological Knowledge Base

The declarative knowledge base of the phonological event parser consists of two types of constraint: phonetic constraints which define well-formedness of phonological events and phonotactic constraints which define well-formedness of syllable events. The phonetic constraints are defined in a temporal constraint logic which is implemented in network descriptions of phonological events. A phonological event `plosive`, for example, is defined according to temporal precedence relations between the acoustic events `occlusion`, `burst` and `noisy`. The syllable event phonotactics is formulated in terms of a network of constraints on overlap and immediate precedence relations between autosegmental phonological events. A permissible onset, for example, given a `voiceless apical plosive` specification at the phonological level for German, requires that the next specification be either `vowellike` or `voiced uvular fricative` (depending on the /r/ variant); a lateral is not permissible in this phonotactic position. The phonotactic constraints are defined in the autosegmental representation with respect to a primary tier which is interpreted as an abstract timimg tier. Each element of the primary tier defines (provides a reference to) constraints on overlap and immediate precedence of phonological events in a particular syllable position. The primary autosegmental tier is represented as a finite state automaton which interprets these constraints. During phonological parsing, the primary tier provides control and top-down constraints for the input acoustic event representation. The phonological knowledge base of the phonological parser contains the complete event-based phonotactics of standard German (?), that is to say, it describes all possible syllables of the language together with a corpus lexicon which allows a distinction to be made between `actual` and `potential` forms. The phonological event parser and the phonological knowledge base have been tested for consistency within the context of logical evaluation (against manually labelled data) and have been shown to achieve a logical recognition rate of 99.7% on these test data.

## 7  Phoneme and Syllable Recognition

The phonological parser analyzes the acoustic event lattice into syllable event structures which are then passed to a morphoprosodic parser. Rather than performing a complete segmentation into connected chart nodes, this approach performs a mapping from the temporal annotations or boundary points to temporal relations between the hypotheses. Parsing is then carried out using relations rather than the temporal annotations. Gaps are described in terms of immediate precedence relations and overlapping hypotheses are described in terms of overlap relations. Analysis is undertaken by a finite automaton labelled with constraints on autosegmental representations, in this case, acoustic event lattices. A well-formed and, in the general case, underspecified representation of the syllable structure is provided in accordance with the constraints. Output to the morphoprosodic parser consists of underspecified phonological event structures which describe classes of phonological segments. The fact that there may be more information in the input acoustic event lattice than is required by the phonology is not a problem for the primary tier finite automaton since it only requires that the constraints specified on the arcs of the network are fulfilled. For the case that there is less information in the acoustic event lattice than is required by the phonological parser, constraint relaxation can be performed by altering the parameter settings of the parser so that unreliable or incorrect information is not weighted as heavily as information which is plays a role in the phonotactic context. Underspecified information can be enhanced (i.e. further specified) by the positional constraints provided by the phonotactic network. In this way, phonological parsing has been made more robust. When the phonological parser was parametrized so that constraints were relaxed completely, the syllable recognition rate on continuous speech data improved from an initial 14.5% to 37%. Phoneme recognition which is a side effect of syllable recognition (i.e. derived top-down from the phonemes which occur in the recognized syllables) improved from 49.6% to 72.5%.

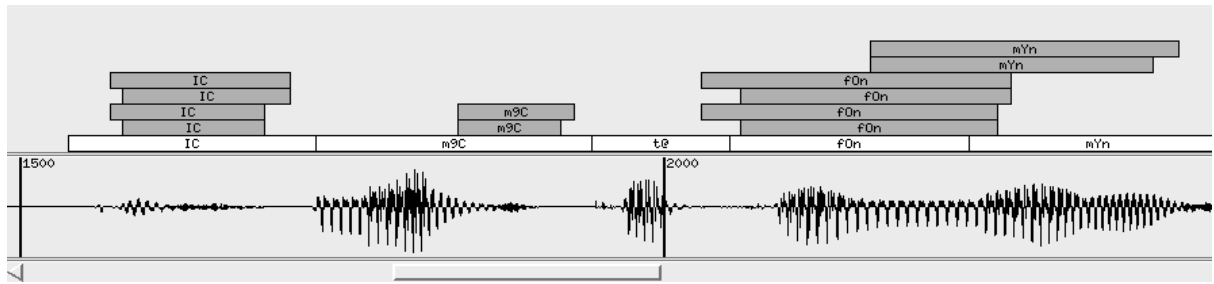

Figure 2: Syllable Output of the Phonological Parser

In Figure ??, the mapping from acoustic to phonological events as generated incrementally by the phonological parser for a section of the utterance token *Ich möchte von München über Nürnberg nach Hamburg fahren* is presented. Figure ?? is a visualization of the syllable evaluation showing the temporally annotated reference path (light shading) and the correctly evaluated syllable hypotheses (dark shading) of the initial part of the utterrance. Other possible syllable hypotheses have been filtered out and are not shown in the visualization but are passed to the morphoprosodic parser for futher analysis. The apparent gaps in the output representation are due to the fact that the temporally annotated syllable hypotheses represent the core area where the phonological constraints are fulfilled. Due to the fact that many phonological constraints have been relaxed, slight differences in the temporal annotations of the same syllable can be observed. When generalisation over these temporal annotations is performed, only one is selected.

## 8 Summary

A new approach to phoneme recognition based on acoustic events was presented. Acoustic events are nonsequential sub–phoneme units which can be mapped to syllables and phonemes by a knowledge–based phonological parser. To demonstrate the feasibility of our approach an acoustic event recognizer and a phonological event parser have been built and tested on a speaker–dependent task. A phoneme recognition rate of 72.5% was obtained.

The fact that the phonological parser outperforms the HMM–based recognition of phonemes from acoustic events shows the potential that lies in applying phonological knowledge in phoneme recognition.

Currently, the use of formant frequencies to further improve recognition rates for events corresponding to places of articulation and vowel quality is investigated. Another area of research is the incorporation of feedback from the phonological parser to the acoustic event recognizer in the framework of incremental and interactive processing of speech signals (?). This would allow deactivating some detectors within certain regions of the signal thus speeding up the recognition process while reducing the rate of false alarms.

At the phonological level, statistical information such as average syllable and phonological event durations and information on the frequency of linguistic items are being incorporated which will allow a specialisation of the knowledge base. Currently, experiments are being carried out in connection with the reliability of events in order to find the optimal parameter settings for the parser with respect to phonological constraint relaxation.